\documentclass{article}
\usepackage[dvipsnames]{xcolor}
\usepackage[preprint]{neurips_2024}  
\usepackage{graphicx}
\usepackage{booktabs}
\usepackage{longtable}
\usepackage{adjustbox}
\usepackage{amsmath, amssymb}
\usepackage{siunitx}
\usepackage{url}
\usepackage{hyperref}
\usepackage{natbib}
\usepackage{microtype}
\usepackage{algorithm}
\usepackage{algorithmic}

\title{Bridging Electronic Health Records and Clinical Texts: Contrastive Learning for Enhanced Clinical Tasks}

\author{
  Sara Ketabi\\
  Vector Institute, University of Toronto, The Hospital for Sick Children
  \And
  Dhanesh Ramachandram \\
  Vector Institute
}

\begin{document}

\maketitle

\begin{abstract}
Conventional machine learning models, particularly tree-based approaches, have demonstrated promising performance across various clinical prediction tasks using electronic health record (EHR) data. Despite their strengths, these models struggle with tasks that require deeper contextual understanding, such as predicting 30-day hospital readmission. This can be primarily due to the limited semantic information available in structured EHR data. To address this limitation, we propose a deep multimodal contrastive learning (CL) framework that aligns the latent representations of structured EHR data with unstructured  discharge summary notes. It works by pulling together paired EHR and text embeddings while pushing apart unpaired ones. Fine-tuning the pretrained EHR encoder extracted from this framework significantly boosts downstream task performance, e.g, a 4.1\% AUROC enhancement over XGBoost for 30-day readmission prediction. Such results demonstrate the effect of integrating domain knowledge from clinical notes into EHR-based pipelines, enabling more accurate and context-aware clinical decision support systems.

\end{abstract}

\section{Introduction}

Electronic health record (EHR) is one of the most common data modalities in healthcare, comprising various patient-related information including demographic records, lab results, and medication history \cite{johnson2023mimic}. Conventional machine learning (ML) algorithms, e.g., XGBoost, have achieved significant performance in extracting useful patterns from this data and using them to predict various clinical tasks \cite{li2024development}. These models have even outperformed deep learning (DL) approaches when applied to EHR data, making them the preferred models in this domain. However, there are several challenging tasks, such as hospital readmission prediction or mortality forecasting, that depend on complex and context-rich factors unavailable in EHR data. In such cases, models trained solely on tabular EHR features usually struggle to capture the nuanced clinical reasoning and latent patterns that human experts rely on, leading to suboptimal performance.

To overcome this challenge and improve the predictive performance of ML models in such complex tasks, additional data modalities representing domain knowledge can be integrated with EHR. A potential option can be clinical texts, which are provided by clinicians and can be invaluable sources of their knowledge. In contrast to non-DL models which are mostly applicable to unimodal tasks, DL architectures can learn joint feature representations of multimodal frameworks, such as EHR and these texts in our case, to enhance the representation of single modalities.

Contrastive learning (CL) \cite{chen2020simple}, as a subtype of self-supervised learning, can learn the alignment between multiple input modalities by minimizing the distance between paired samples while maximizing that between unpaired modalities. Through this process, the model learns to generate meaningful representations, where semantically related elements, such as EHR and its corresponding text, are embedded close to each other. This alignment enhances the model's ability to perform downstream tasks such as classification and multimodal reasoning. 

Prior works have proposed multimodal CL frameworks in the medical domain, mostly on medical images and radiology reports \cite{huang2021gloria,wang2022multi,ji2021improving}, to improve the performance of downstream clinical tasks. While these approaches have demonstrated significant potential in  image-text settings, there has been relatively limited exploration of CL applied to other clinical modalities. In particular, the integration of EHR with unstructured textual data, such as discharge summaries, remains less explored. 

While some recent studies have investigated CL using EHR data and clinical text \cite{ma2024global,king2023multimodal}, they mainly treat EHR as temporal sequences (e.g., visit-based or longitudinal records) paired with textual summaries. Such approaches can capture dynamic patient trends over time. In contrast, our work utilizes static EHR data along with corresponding discharge summaries. This distinction is important in cases where time series elements are unavailable or unnecessary. By working with static EHR and clinical notes, our framework increases the applicability of multimodal CL in clinical settings and serves as a foundation model for downstream tasks operating on more readily available forms of EHR data.

To that end, we propose a multimodal CL framework trained on static EHR data and unstructured discharge summary notes to enrich the semantic content of EHR representations. By aligning the representations of these two modalities in a shared embedding space, our framework can capture clinically meaningful patterns that may not be apparent from structured EHR alone. Subsequently, we fine-tune the learned representations on two downstream classification tasks, which conventional models such as XGBoost achieve low predictive performance on, to measure the effectiveness of our framework in enhancing these tasks. 




\section{Literature Review}
\label{Lit}
\subsection{Deep Learning Architectures for Tabular Data Processing}

Classical ML models, such as XGBoost and Random Forest, have achieved promising results on various tabular data classification problems, particularly in the clinical domain. These problems include, but are not limited to, kidney
injury risk forecasting \cite{li2025predicting}, infection diagnosis \cite{horng2017creating}, and drillium prediction \cite{ma2024machine}. Efficiency and ease of implementation have made these models a popular and reliable choice for practitioners working with structured tabular data.

Nevertheless, to address the limitations of these models, e.g., their inability to learn hierarchical feature representations or handle multimodal inputs, different DL architectures have been proposed to process tabular data more effectively. TabNet \cite{arik2021tabnet}, is a DL algorithm consisting of an attentive feature selection mechanism, which can determine the importance of individual features and select the most informative ones at each decision step. Neural Oblivious Decision Ensembles (NODE) \cite{popov1909neural} combines tree-based models with neural networks by applying backpropagation to a differentiable ensemble of decision trees. Tabtransformer \cite{huang2020tabtransformer}involves a sequence of attention-based transformer layers to convert original feature embeddings into useful contextual ones. 


Despite these advances, consistent improvement of DL models over classical ML has not been widely demonstrated in tabular data encoding. Moreover, several complex clinical tasks, e.g., treatment response or long-term outcome prediction, are challenging to tackle using tabular data alone. A potential method for addressing such challenges can be extracting semantic context from other data modalities, such as unstructured clinical notes, by leveraging the multimodal learning capabilities of DL frameworks. Such frameworks can integrate EHR data with free-text input, enabling more semantically rich EHR representations.



\subsection{Multimodal Contrastive Learning Frameworks}

Multimodal CL frameworks have achieved remarkable success in integrating various modalities of input data and learning semantically aligned representations. In the healthcare domain, these approaches have been widely applied to medical images, e.g., chest X-ray, and corresponding radiology reports, aligning the representations of paired images and texts \cite{huang2021gloria,wang2022multi,ji2021improving}. Regarding other modalities, CL has been explored to learn the alignment between tabular data, e.g, Morphometric features, and cardiac magnetic resonance (MR) images \cite{hager2023best}. However, very limited works have proposed CL frameworks for aligning EHR data and clinical texts \cite{ma2024global,king2023multimodal}. Such methods typically assume that EHR data is structured as time series, limiting their applicability in settings where temporal dynamics of patient records, e.g., event timestamps, longitudinal measurements, are unavailable, unreliable, or inconsistently gathered across institutions.

To the best of our knowledge, this is the first study that proposes a CL framework for aligning the representations of static-form EHR data, which is more readily available compared to time series, and clinical texts. This eliminates the dependency on time-series representations and and broadens the applicability of this framework to situations where EHR data is only available in the static format, e.g., hospital admission or discharge summaries. Consequently, the framework can be applied to improve the performance of a wide range of downstream tasks involving the static format of EHR data and across diverse clinical environments.


\section{Methodology}

Our approach in this study can be divided into two main stages: 1) EHR-text contrastive learning (Subsection \ref{cl}), where we pretrained a CL framework to align the representations of static-form EHR data and associated discharge summary notes using a contrastive objective, and 2) Downstream classification fine-tuning (Subsection \ref{DS}), where we adapted the pretrained EHR representations to two classification tasks using labeled data.

\subsection{EHR-Text Contrastive Learning}
\label{cl}
\begin{figure}
\centering
\includegraphics[width=1\textwidth, height=14cm]{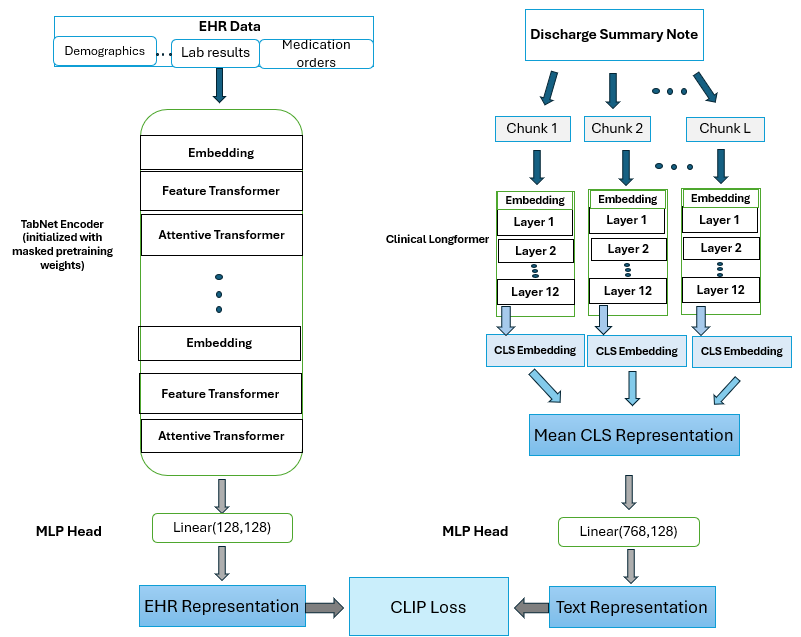} 
\caption{The Proposed EHR-Text CL Framework: EHR is encoded with TabNet initialized with masked pretraining and then forwarded to a linear layer to get its representation. Each discharge summary is divided into chunks of  256 tokens, each encoded by a separate Longformer transformer. The mean of the CLS embeddings extracted from the last layer of all longformer models is then passed through a linear layer to obtain the text representation. A contrastive loss (CLIP-style) is applied between the EHR and text representations to align the related pairs and train the framework.}
\label{CL}
\end{figure}

Our proposed CL framework is designed to align static-form EHR data with unstructured discharge summary notes, leveraging the inherent cross-modal correspondence between these modalities. We trained this framework on EHR-report pairs from the same patient admission, with the objective of minimizing the distance between the embeddings of matched pairs and maximizing the distance between mismatched ones. Consequently, the model learns to embed semantically related EHR and text representations closely in the shared latent space, thereby enriching EHR representations with the contextual and semantic information available in clinical texts. This framework is visualized in Figure \ref{CL}.

To encode the structured EHR inputs $X_{ehr}\in \mathbb{R}^{d}$, where $d$ is the number of features, and get EHR representations, we employed the encoder portion of TabNet. As discussed in Section \ref{Lit},  TabNet utilizes a sequential attention mechanism and sparse feature selection to extract complex interactions within the tabular data. Let $e$ be the EHR representation and $T$ indicate an instance of TabNet:

\begin{equation}
e = T(X_{ehr}) \in \mathbb{R}^{128}
\end{equation}

Following the pretraining approach proposed in the original TabNet paper \cite{arik2021tabnet}, we first pretrained the EHR encoder using a self-supervised masked feature reconstruction task. This pretraining stage enables the model to capture internal dependencies and useful patterns within the EHR modality.

To enhance the training stability and efficiency, we froze the initial embedding and feature-splitting layers of TabNet. These layers are believed to capture general low-level patterns that remain stable across a wide range of tasks. Consequently, the model could focus on adapting deeper layers which can be more abstract and task-specific, significantly reducing the number of trainable parameters.

For the discharge summary notes, we used Longformer \cite{beltagy2020longformer}, a state-of-the-art transformer capable of processing long input sequences (up to 4096 tokens), as the text encoder since it was crucial for capturing context in long discharge documents. We initialized this transformer with ``clinical longformer'' \cite{li2022clinical} weights, a version of Longformer pretrained on MIMIC-III clinical notes. This would enable the model to learn specific representations of clinical terms, enhancing the training process of the main CL framework. Similar to the EHR encoder, we froze the initial layers of the Longformer model, i.e., first 10 layers, and fine-tuned the final two layers to reduce the computational complexity.

As the discharge notes were typically longer than the maximum token size that can be processed by Longformer (4096), we divided each note into chunks of 256 tokens, 
and each chunk was passed independently through the model. Therefore, given L, as an instance of Longformer, and a discharge note $X_{n} = [c_{1},c_{2},...,c_{l}]$, where  $\{c_{i}\}_{i=1}^l$ indicates a separate chunk, and $l$ is the total number of chunks extracted from the text:

\begin{equation}
{\text{t}} = \frac{1}{l}\sum_{i=1}^{l} L(c_{i})[0] 
\in \mathbb{R}^{768}
\end{equation}

The representation of each text sequence can be obtained by using mean pooling over the [cls] token embeddings of chunk representations extracted from the last Longformer layer, which appears as the first element. This embedding can be considered as the aggregation of the whole sequence's representation.

To align the shape of the EHR and text representations and map them onto a shared embedding space, we added a multi-layer perceptron (MLP) projection head to the encoders As shown by \cite{chen2020simple}. This module consist of linear layers with sizes (768, 128) and (128,128), applied to the text and EHR representations, respectively.



The framework was trained using a CLIP loss \cite{radford2021learning}. Given a batch of $N$ paired EHR--text samples $\{(e_i, t_i)\}_{i=1}^N$, let $e_i$ and $t_i$ be the normalized representations of the structured EHR and unstructured discharge summary notes, respectively. The cosine similarity between each EHR--text representation pair can be calculated as:

\[
s_{ij} = \text{sim}(e_i, t_j) = \frac{e_i^\top t_j}{\|e_i\| \, \|t_j\|}
\]

The EHR-to-Text loss is defined as:

\[
\mathcal{L}_{\text{ehr} \rightarrow \text{text}} = \frac{1}{N} \sum_{i=1}^{N} -\log \left( \frac{\exp(s_{ii}/\tau)}{\sum_{j=1}^{N} \exp(s_{ij}/\tau)} \right)
\]

Similarly, the Text-to-EHR loss is:

\[
\mathcal{L}_{\text{text} \rightarrow \text{ehr}} = \frac{1}{N} \sum_{i=1}^{N} -\log \left( \frac{\exp(s_{ii}/\tau)}{\sum_{j=1}^{N} \exp(s_{ji}/\tau)} \right)
\]
$\tau$ is the temperature parameter \cite{radford2021learning} used for controlling the sharpness of the similarity distribution.

The final CLIP loss is the summation of the two directions:

\[
\mathcal{L}_{\text{CLIP}} =  \mathcal{L}_{\text{ehr} \rightarrow \text{text}} + \mathcal{L}_{\text{text} \rightarrow \text{ehr}} 
\]

\subsection{Downstream Tasks}
\label{DS}

\begin{figure}
\centering
\includegraphics[width=0.7\textwidth, height=14cm]{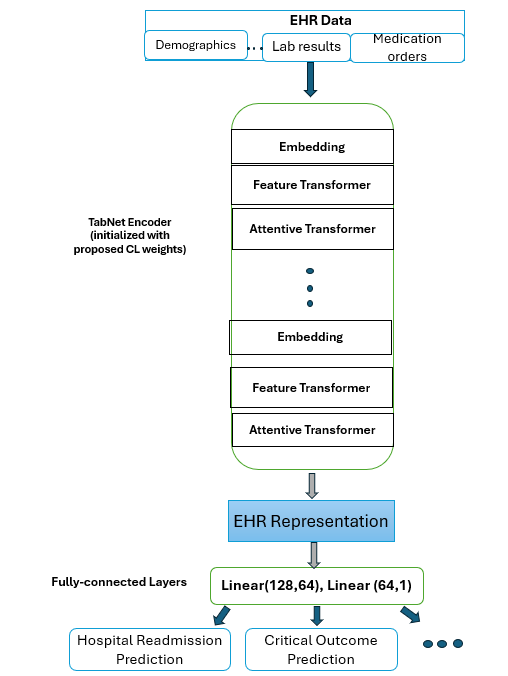} 
\caption{The Downstream Classification Architecture: The TabNet encoder is initialized with the proposed CL weights. The initial embedding and feature-splitting layers are kept frozen, and the remaining layers along with two fully-connected layers are fine-tuned for each downstream task.}
\label{ds}
\end{figure}

Following the pretraining of our EHR-text CL framework, we evaluated the effectiveness of the learned EHR representations by fine-tuning them on two clinically significant downstream classification tasks. Notably, only the EHR encoder was used during fine-tuning, assessing the impact of transferring cross-modal semantic knowledge from clinical notes into structured EHR representations. The architecture used for fine-tuning the downstream tasks can be observed in Figure \ref{ds}. 

The first downstream task is 30-day hospital readmission prediction, where the objective is to determine whether a patient is likely to be readmitted within 30 days of discharge, a key factor determining in patient health monitoring and hospital resource management. The second task is critical outcome prediction, which involves identifying patients at risk of life-threatening events such as hospital mortality.

For both tasks, we employed the TabNet encoder architecture as our classifier. We initialized the model using the pretrained weights obtained from the CL stage. To preserve the general low-level features acquired during pretraining, we froze the embedding and initial feature-splitting layers and fine-tuned only the remaining layers along with two additional linear layers on the downstream tasks.

The models were fine-tuned using the binary cross-entropy loss, as indicated in Equation \ref{eq5}:

\begin{equation}
\label{eq5}
    \mathcal{L}_{\text{FT}} = - \frac{1}{N} \sum_{i=1}^{N} \left[ y_i \log \hat{y}_i + (1 - y_i) \log (1 - \hat{y}_i) \right]
\end{equation}

Where $y_i \in \mathbb [0,1]$ is the ground-truth downstream label corresponding to the $i^{th}$ datapoint, and $\hat{y}_{i}$ is the predicted probability for that datapoint.

\section{Experiments and Results}

\subsection{Dataset}

For training the proposed CL framework, we utilized MIMIC-IV \cite{johnson2023mimic}, an open-source dataset accessed under the associated data use agreement with online consent. This dataset contains EHR information related to  admissions to Beth Israel Deaconess Medical Center between 2008 and 2019. It consists of structured data such as demographic records, medication orders, lab results, and unstructured clinical texts, e.g., radiology reports and discharge summary notes. 

Since our CL framework requires paired EHR data and discharge summary notes, we excluded all cases for which these notes were not available, resulting in a total of 376,021 EHR-text pairs and 105 features (columns) used for training.

To prepare the EHR data for training, we first removed redundant or non-informative columns, such as patient and admission ID, which do not contribute meaningful clinical information. We also excluded features containing missing values and any columns directly related to the downstream classification targets, e.g., readmission outcome, to avoid data leakage. This resulted in a total of 105 features used for training our CL framework.
We determined all features with categorical values or numerical variables with fewer than three unique values as categorical features. The remaining features, showing a wider numeric range, were highlighted as numerical. Prior to being processed by the model, categorical features were ordinally encoded. The TabNet model then employs an embedding layer to transform these encoded values into dense vectorized representations, enabling the model to capture semantic relationships between feature categories. 

For the text data, we standardized all discharge summaries by converting them into lowercase. We also removed all dates, numbers, additional spaces, and punctuation marks. Furthermore, several sections within the discharge summaries which do not contain important clinical information, such as Technique, Discharge Instructions, and administrative segments, were excluded.

To fine-tune the CL framework on downstream classification tasks, we employed a held-out portion of the dataset which was excluded during the pretraining CL stage, i.e., datapoints without corresponding discharge summary notes. This design ensures no data leakage from the pretraining. We randomly sampled five non-overlapping subsets from this data to robustly evaluate model performance. Each subset consisted of 6,000 datapoints for model training and validation, and 2,496 datapoints were utilized for final testing.

\subsection{Implementation Details}

For training the proposed CL framework, we applied Distributed Data Parallel (DDP) computing to accelerate the training process. The experiments were conducted on four ``NVIDIA A40'' GPUs. The model was developed using \textit{AdamW} optimizer with a learning rate of 1e-4 and weight-decay of 1e-4 to regularize the model and reduce overfitting. The batch-size and number of epochs were set to 64 and 13, respectively. The temperature in the CLIP loss was chosen as 0.1.

For the downstream classification fine-tuning, DDP was employed across 4 ``NVIDIA RTX6000'' GPUs. AdamW with an increased learning rate of 5e-4 and weight-decay of 1e-4 was set as the optimizer. The batch-size remained as 64, and the models were fine-tuned for 15 epochs.

The code used for developing the experiments and the pretrained framework's weights are provided in the supplementary material.

\subsection{Evaluation Setting}

We evaluated the effectiveness of the proposed EHR-text CL framework by comparing the performance of a TabNet model initialized with the pretrained CL weights and fine-tuned on the aforementioned downstream classification tasks with two baseline models:
\begin{itemize}
\item XGBoost
\item A TabNet model with the same architecture but initialized with only TabNet masked pretraining weights without proposed CL pretraining
\end{itemize}

We used the Area Under the Receiver Operating Characteristic Curve (AUC) as our primary performance metric, as it is well-suited for imbalanced classification tasks and can reflect the trade-off between sensitivity and specificity. We computed the mean and standard deviation across the five random seeds to compare the models and evaluate their generalization.

Moreover, we aimed to assess the impact of the proposed framework on the amount of training data required for effective fine-tuning. This was achieved by reducing the number of training datapoints by 50\% and re-evaluating the performance of all three models on the reduced dataset.

\subsection{Results}

Tables \ref{table1} and \ref{table2} demonstrate the mean test AUC achieved by the proposed CL-based fine-tuned model as well as the baselines for 30-day readmission and critical outcome prediction, respectively. Moreover, to assess the statistical significance of performance differences, we conducted two-tailed t-tests on the AUC scores obtained across the five seeds. The detailed results of these tests are presented in the supplementary material.

\subsection{Hospital Readmission Prediction}

 \begin{table}[h]
  \centering 
  \caption{Mean Test AUC Results for 30-day Hospital Readmission Prediction (When 100\% and 50\% of training data are used for model fine-tuning)} 
  
  \begin{tabular}{lll}
  \toprule
    \textbf{Model} & \textbf{100\% Training Data} & \textbf{50\% Training Data} \\
    \midrule
    XGBoost & 0.777 ($\pm0.012$) & 0.756 ($\pm0.018$)\\ 
    TabNet (initialized with masked pretraining)  & 0.759 ($\pm0.025$) & 0.741 ($\pm0.016$)\\ 
    TabNet (initialized With proposed CL) &  \textbf{0.809} ($ \pm0.016$) & \textbf{0.788} ($ \pm0.008$)\\   
    \bottomrule
  \end{tabular}
  \label{table1} 
\end{table}

As can be observed in Table \ref{table1}, the proposed model, achieving a mean test AUC of 0.809 on the whole dataset,  consistently outperforms both XGBoost and the masked-pretrained TabNet by 4.12\% (p-value=0.0082, statistically significant) and 6.59\% (p-value=0.0074, statistically significant), respectively. Moreover, when only 50\% of the training data, i.e, 3000 datapoints, is used for model fine-tuning, this model still demonstrates strong discriminative capability, i.e., an AUC of 0.788, which is higher than the performance of both baselines even when trained on the whole training data.



\subsection{Critical Outcome Prediction}

\begin{table}[h]
  \centering 
  \caption{Mean Test AUC Results for Critical Outcome Prediction (When 100\% and 50\% of training data are used for model fine-tuning)} 

  \begin{tabular}{lll}
  \toprule
    \textbf{Model} & \textbf{100\% Training Data} & \textbf{50\% Training Data} \\
    \midrule
    XGBoost &  0.725($\pm 0.017$) & 0.725($\pm 0.048$)\\ 
    TabNet (initialized with masked pretraining)  &  0.745 ($\pm 0.019$) & 0.693 ($\pm 0.014$)\\ 
    TabNet (initialized With proposed CL) &  \textbf{0.821} ($ \pm 0.022$) & \textbf{0.782} ($ \pm 0.027$)\\   
    \bottomrule
  \end{tabular}
  \label{table2} 
\end{table}
As indicated in Table \ref{table2}, the performance gap is even larger for critical outcome prediction, where the CL-based fine-tuned model surpasses XGBoost and the masked-pretrained model by 13.25\% (p-value=0.0001, statistically significant) and 10.2\% (p-value=0.0004, statistically significant), respectively, on the whole training data. Moreover, this model still shows superior performance compared to both baselines when only 50\% of the training data is utilized.

\section{Discussion}

In this study, we proposed a novel multimodal CL framework on EHR and discharge summary notes, to align the representations of the two modalities and learn rich and task-agnostic EHR representations. By jointly pretraining these modalities using a CLIP loss, our approach enables the EHR encoder to benefit from the clinical details and semantic information available in discharge summaries. Fine-tuning this framework on two important downstream clinical prediction tasks, namely 30-day hospital readmission and critical outcome prediction demonstrates that CL remarkably improves classification performance over conventional baselines.

Our results indicate that the TabNet model initialized with the proposed CL-pretrained weights consistently outperforms both a TabNet pretrained with only masked modeling weights and a powerful non-DL baseline, i.e, XGBoost, across five random seeds of data split. The evaluation was performed based on the average test AUC, which is a widely used metric in clinical settings. These findings suggest that aligning static-form EHR data with semantically rich discharge summaries during pretraining leads to a more meaningful initialization for tabular-based downstream predictive tasks, effectively guiding the model toward clinically relevant EHR representations.


TabNet’s use of sequential decision steps and attentive feature selection shows to be highly compatible with contrastive pretraining. The model not only learns generalizable feature representations but also is able to incorporate new semantic information efficiently during fine-tuning. Freezing the embedding and initial splitting layers during downstream fine-tuning helped retain generalized feature patterns learned through pretraining, while allowing the deeper layers to adjust to the target task.

The framework’s strong performance on readmission and critical outcome prediction, two tasks with high clinical importance, without relying on text during fine-tuning or inference, demonstrates the practical applicability of multimodal contrastive  pretraining in real-world hospital settings. Moreover, our model architecture is modular and modality-agnostic. Therefore, it can be adapted to other types of structured data, e.g., imaging metadata, or unstructured clinical texts, e.g., radiology reports, showing its generalizability. Furthermore, under the low-resource regime, when the framework was fine-tuned on only half of the training data, it could outperform both baselines trained on the whole data across both downstream tasks, suggesting that it learns more generalizable features regardless of the data size. This finding supports the hypothesis that CL-based pretraining acts as a strong regularizer and can be particularly advantageous when labeled data is limited.



Despite the promising performance achieved by the proposed framework, there are several limitations associated with this work. First, the reliance on discharge summaries during pretraining restricts the model to cases where high-quality clinical notes are available. This may introduce selection bias, as patients without notes may systematically differ in clinically relevant ways. Second, although TabNet supports interpretability, our current analysis did not explore how contrastive pretraining affects the learned feature importance. Moreover, our evaluation focuses on two binary classification tasks. We should also investigate whether the proposed framework generalizes well to other settings, such as multi-label classification, or generative modeling, e.g., report generation from structured EHR data.

Several steps could be performed for extending this study in future work. First, we plan to evaluate the framework on another clinical dataset to assess its cross-domain generalizability. Second, incorporating additional modalities—such as imaging (e.g., chest X-rays), into the multimodal framework could lead to even richer data representations. Furthermore, alternative contrastive objectives, such as Triplet Loss with hard negative mining or supervised CL, could be explored to determine whether the loss choice could have a significant impact on the results. Moreover, we will explore TabNet interpretability and attention weight analysis to identify whether the pretrained model attends to different clinical variables than a randomly initialized one. Finally, we will use generative language models to generate discharge summary notes from EHR data, an important step towards explainability improvement of tabular-based models.




\section{Conclusion}

In this work, we introduced a deep multimodal CL framework that leverages paired static-form EHR data and discharge summary notes to learn semantically rich EHR representations. Our results indicate that fine-tuning this framework on two high-impact clinical downstream tasks, i.e., 30-day hospital readmission and critical outcome prediction, outperforms both a non-CL-based pretrained and conventional ML baselines based on test AUC. Moreover, our downstream model works completely independent of clinical texts, providing a foundation for a wide range of tabular-based clinical analysis tasks. Deploying our model in real-world clinical settings  can lead to more effective decision support systems, thereby enhancing clinical workflow and patient outcomes.

\end{document}